\documentclass[a4paper,twoside]{article}

\usepackage{epsfig}
\usepackage{subcaption}
\usepackage{calc}
\usepackage{amssymb}
\usepackage{amstext}
\usepackage{amsmath}
\usepackage{amsthm}
\usepackage{multicol}
\usepackage{pslatex}

\usepackage{natbib}

\usepackage{algorithm2e}
\usepackage[bottom]{footmisc}

\usepackage{url}
\usepackage{amsmath}
\usepackage{multirow}
\usepackage{graphicx}
\usepackage{xcolor}

\usepackage[colorlinks=true, allcolors=blue, citecolor=blue]{hyperref}

\usepackage{SCITEPRESS}     % Please add other packages that you may need BEFORE the SCITEPRESS.sty package.

\begin{document}

\title{On Few-Shot Prompting for Controllable Question-Answer Generation in Narrative Comprehension\thanks{Preprint - Accepted for publication at CSEDU 2024.}
}

\author{\authorname{Bernardo Leite\sup{1,2}\orcidAuthor{0000-0002-9054-9501} and Henrique Lopes Cardoso\sup{1,2}\orcidAuthor{0000-0003-1252-7515}}
\affiliation{\sup{1}Faculty of Engineering of the University of Porto (FEUP)}
\affiliation{\sup{2}Artificial Intelligence and Computer Science Laboratory (LIACC)}
\email{\{bernardo.leite, hlc\}@fe.up.pt}
}


\abstract{Question Generation aims to automatically generate questions based on a given input provided as context. A controllable question generation scheme focuses on generating questions with specific attributes, allowing better control. In this study, we propose a few-shot prompting strategy for controlling the generation of question-answer pairs from children's narrative texts. We aim to control two attributes: the question's explicitness and underlying narrative elements. With empirical evaluation, we show the effectiveness of controlling the generation process by employing few-shot prompting side by side with a reference model. Our experiments highlight instances where the few-shot strategy surpasses the reference model, particularly in scenarios such as semantic closeness evaluation and the diversity and coherency of question-answer pairs. However, these improvements are not always statistically significant. The code is publicly available at \url{github.com/bernardoleite/few-shot-prompting-qg-control}.}

\onecolumn \maketitle \normalsize \setcounter{footnote}{0} \vfill

\section{Introduction}
% QG task extensive definition
The task of Question Generation (QG) involves the automatic generation of well-structured and meaningful questions from diverse data sources, such as free text or knowledge bases \citep{rus2008question}.
% Controllability in QG
Controllable Question Generation (CQG) holds significant importance in the educational field \citep{kurdi_2020_education}, as it boosts the creation of customized questions tailored to student's specific needs and learning objectives.

\begin{figure}[htp]
    \centering
    \includegraphics[scale=0.35]{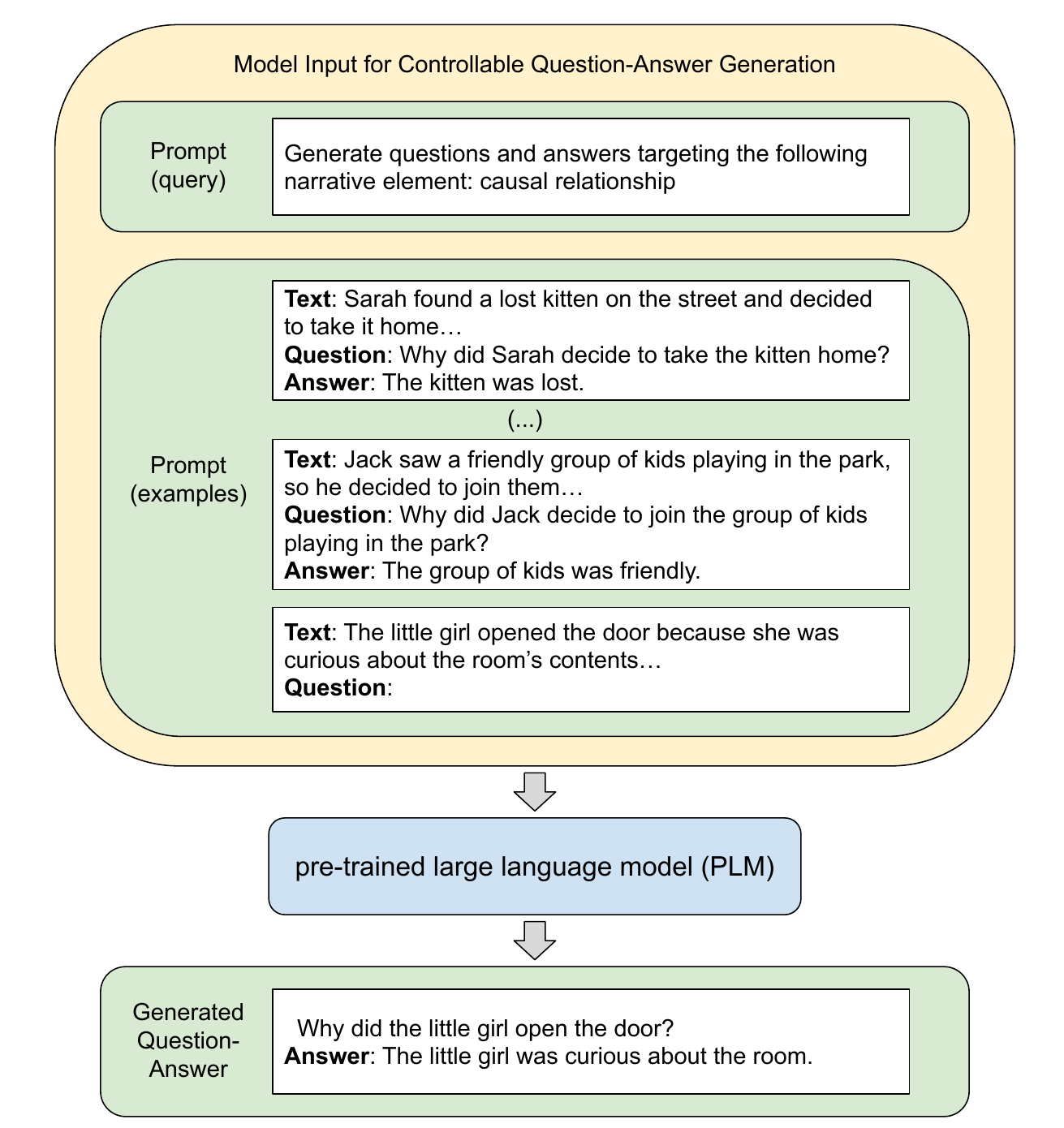}
    \caption{A simplistic example of few-shot prompting for controllable question-answer generation.} 
    \label{fig:pipeline}
\end{figure}

% General method for QG and where they fail
From a methodological perspective, some prior works on QG have focused on fine-tuning large pre-trained language models (PLM) \citep{zhang_2021_qg_survey} for generating questions (output) given a source text and possibly a target answer (input). This is also the case for CQG, with the addition of incorporating controllability labels into the input to serve as guidance attributes during the generation process \citep{zhao_2022_cqg_acl,ghanem_2022_cqg_acl}. After undergoing fine-tuning, the models have demonstrated good performance \citep{ushio_2022_qg_t5}. However, utilizing these models demands a custom model design (e.g., architecture, choice of hyperparameters) and substantial computational resources for training, which may deter practitioners who prefer a convenient ``plug-and-play'' AI-assisted approach, enabling them to effortlessly interact with a QG system without the need to undergo model design and training \citep{wang_2022_towards_aied}.

% GPT-3 and prompts technique, proposal
Motivated by this, we explore a few-shot prompting strategy to address the CQG task in the context of narrative comprehension. We aim to control the generation of question-answer pairs conditioned by their underlying narrative elements (e.g., character, setting, acting) and question explicitness (explicit or implicit). We explore the prompting paradigm, where a prompt specifies the desired generation task. Figure~\ref{fig:pipeline} shows an illustrative example of the few-shot prompting strategy for CQG. Prompting offers a straightforward interface and a significant level of control to interact with PLMs and tailor them according to various generation requirements. Its practicality and simplicity have contributed to its popularity as a means of customizing PLMs for diverse tasks \citep{wang_2022_towards_aied,liu_2023_prompt}. 

% Side Goal
We assess the few-shot strategy's performance alongside a smaller yet reference model, where fine-tuning is applied. To this end, we empirically evaluate the question-answer pairs generated by the two methods using similarity and quality metrics.

% Clarification about primary goal
Nonetheless, we stress out that our primary goal is not to determine whether the few-shot strategy is better or worse than fine-tuning for question generation control. Such an assessment would necessitate a separate, comprehensive study focused on an in-depth model comparison. Instead, our primary aim is to delve into the potential of a few-shot strategy for controlled question-answer generation. We indeed build and incorporate, for reference, the results obtained by a fine-tuned model based on previous findings \citep{zhao_2022_cqg_acl,ghanem_2022_cqg_acl}, known for achieving commendable results in CQG.

% Contributions
In summary, our primary contribution is a few-shot strategy, based on the prompting paradigm, for controlling the generation of question-answer pairs based on their narrative elements, explicitness, or both. While using prompts via few-shot prompting has been explored in previous research, the novelty of our study lies in its focused application on \textit{narrative comprehension} through the controlled generation of \textit{both} questions and answers. As a result, our analysis contributes to the ongoing discourse by providing valuable and unique insights into this unexplored area.

\section{Background and Related Work}
% Terms clarification
\textbf{Few-Shot}: This study presents a strategy for CQG based on few-shot prompting. To ensure clarity in terminology, we provide definitions for few-shot and fine-tuning as elucidated by \citet{brown_2020_terms}. Few-Shot pertains to approaches where the model receives limited task demonstrations as conditioning context during inference \citep{radford_2019_gpt2}, without allowing weight updates. Fine-Tuning refers to the process of adjusting the weights of a pre-trained model by training it on a dataset tailored to a specific task. This involves utilizing a significant number of labeled examples.

% prior research -- CQG for EDU, via fine-tuning
\textbf{Controllable Question Generation (CQG)}: Prior research has explored CQG for education. \citet{ghanem_2022_cqg_acl} employed fine-tuning with the T5 model~\citep{raffel_2020_t5} to control the reading comprehension skills necessary for formulating questions, such as understanding figurative language and summarization. Similarly, \citet{zhao_2022_cqg_acl} aimed to control the generated questions' narrative aspects. Still, through fine-tuning, \citet{leite_2023_cqg_aied} propose to control question explicitness using the T5 model. 
% prior research -- CQG for EDU, via few-shot
Finally, via few-shot, \citet{elkins_2023_cqg_aied} propose to address the task of CQG by controlling three difficulty levels and Bloom's question taxonomy \citep{bloom_2002_revised} for the domains of machine learning and biology.

% our work
In our study, we make use of the FairytaleQA \citep{xu_2022_fairytaleqa} dataset, which consists of question-answer pairs extracted from stories suitable for children. This dataset has been investigated by two studies above mentioned \citep{zhao_2022_cqg_acl,leite_2023_cqg_aied} for CQG via fine-tuning. 

% novelty/contributions?
To the best of our knowledge, this is the first study addressing few-shot prompting for controlling the generation of both questions and answers in the narrative comprehension domain.

\section{Purpose of QG Control Elements} \label{sec:purpose_qas}

We chose FairytaleQA dataset \citep{xu_2022_fairytaleqa} because its texts and the corresponding question-answer pairs align with the goal of \textit{supporting narrative comprehension}. As highlighted by \citet{xu_2022_fairytaleqa}, narrative comprehension is a high-level skill that strongly correlates with reading success \citep{lynch_2008_reading}.
Additionally, narrative stories possess a well-defined structure comprising distinct elements and their relationships. This dataset stands out since education experts have annotated each question, following evidence-based narrative comprehension
frameworks \citep{paris_2003_narrative,alonzo_2009_narrative}, and addressing two key attributes: \textit{narrative elements} and \textit{explicitness}. Narrative elements we aim to control are as follows:
\begin{itemize}
    \item \textbf{Character}: These require test takers to identify or describe the characteristics of story characters.
    \item \textbf{Setting}: These inquire about the place or time where story events occur and typically start with ``Where'' or ``When''.
    \item \textbf{Action}: These focus on the behaviors of characters or seek information about their actions.
    \item \textbf{Feeling}: These explore the emotional status or reactions of characters, often framed as ``How did/does/do...feel.''.
    \item \textbf{Causal relationship}: These examine the cause-and-effect relationships between two events, often starting with ``Why'' or ``What made/makes''.
    \item \textbf{Outcome resolution}: These ask for the events that result from prior actions in the story, typically phrased as ``What happened/happens/has happened...after...''.
    \item \textbf{Prediction}: These request predictions about the unknown outcome of a particular event based on existing information in the text.
\end{itemize}
Question explicitness is defined as follows:
\begin{itemize}
    \item \textbf{Explicit}: The answers are directly present in the text and can be located within specific passages.
    \item \textbf{Implicit}: Answers cannot be directly pinpointed in the text, requiring the ability to summarize and make inferences based on implicit information.
\end{itemize}

\section{Method}

\subsection{Few-Shot Prompting for CQG}

Let $E$ be a example set containing $K$ text passages and their corresponding question-answer pairs, denoted as $E = \{(x_i, y_i)\}_{i=1}^K$, where $x_{i}$ represents the text passage and $y_{i}$ represents the associated question-answer pair.

The few-shot prompting process can be represented as follows: Given a $query$, the example set $E$ and a new text passage $x_{new}$, the aim is to generate a question-answer pair ($q_{new}, a_{new}$). This can be formulated as:
\begin{equation}
(q_{\text{new}}, a_{\text{new}}) = \text{{PLM}}(query, E, x_{\text{new}}),
\end{equation}
where PLM represents the pre-trained language model that generates the question-answer pair ($q_{new}, a_{new}$) based on a $query$, the example set $E$, and the new text passage $x_{new}$.
$query$ is the textual instruction designed to control the generation of question-answer pairs conditioned to the desired attributes. In this study, it can assume four formats\footnote{See Figures \ref{fig:cqg_nar} and \ref{fig:cqg_ex} for concrete examples.}:

\begin{enumerate}
  \item No Control (baseline): ``\textit{Generate questions and answers from text}:''
  \item Narrative Control: ``\textit{Generate questions and answers targeting the following narrative element: \textsc{$\langle$nar$\rangle$}}:''
  \item Explicitness Control: ``\textit{Generate \textsc{$\langle$ex$\rangle$} questions and answers}:''
  \item Narrative + Explicitness Control: ``\textit{Generate \textsc{$\langle$ex$\rangle$} questions and answers targeting the following narrative element: \textsc{$\langle$nar$\rangle$}}:''
\end{enumerate}
The concrete textual form of the query has been obtained from preliminary and empirical experimentation.

\subsection{Reference Model for CQG}

\begin{figure}[htp]
    \centering
    \includegraphics[scale=0.45]{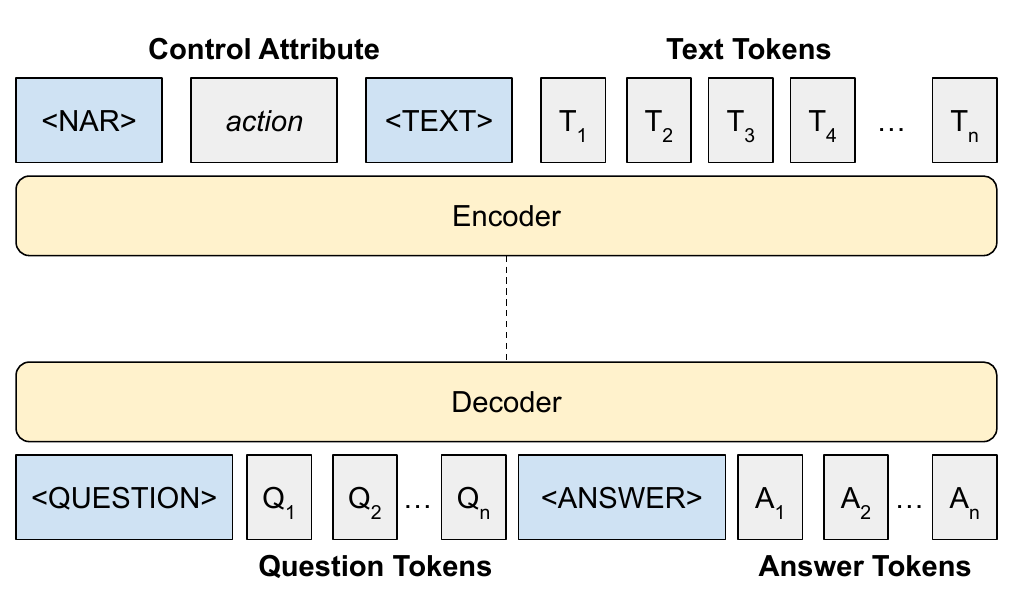}
    \caption{Reference model setup for performing controllable question-answer generation.} 
    \label{fig:fine_tuning_model}
\end{figure}

In the case of the reference model, we frame the CQG task as an encoder-decoder model, where the encoder receives the input text and encodes it into a fixed-length representation known as a context vector. The decoder takes the context vector and generates the output text. Here, the control attributes \textsc{$\langle$nar$\rangle$} or \textsc{$\langle$ex$\rangle$} are added at the start of the input, preceding the section text. Then, the decoder is equipped with labels that serve to differentiate the \textsc{$\langle$question$\rangle$} and \textsc{$\langle$answer$\rangle$} sections of the output. The idea is to guide the model to generate a question-answer pair of the intended type. Figure~\ref{fig:fine_tuning_model} illustrates the reference model setup for generating a question-answer pair targeting the \textit{action} narrative element. This technique is based on recent studies \citep{zhao_2022_cqg_acl,ghanem_2022_cqg_acl} aiming at controlling QG conditioned on specific attributes.

\section{Experimental Setup}

\subsection{FairytaleQA} 
% dataset general info
We make use of FairytaleQA \citep{xu_2022_fairytaleqa}, a dataset composed of 10,580 question-answer pairs manually created by educational experts based on 278 stories. Every question is associated with one of the following narrative elements: \textit{character}, \textit{setting}, \textit{action}, \textit{feeling}, \textit{causal relationship}, \textit{outcome resolution}, or \textit{prediction}. Additionally, each question is accompanied by an explicitness attribute, denoting whether it is explicit or implicit (recall Section~\ref{sec:purpose_qas} for details).
% Stats
Each story consists of approximately 15 sections, each with an average of 3 questions. We use the original train/validation/test splits\footnote{\url{https://github.com/WorkInTheDark/FairytaleQA_Dataset/tree/main/FairytaleQA_Dataset/split_for_training}}, comprising 8,548/1,025/1,007 question-answer pairs, respectively.

\subsection{Data Preparation} \label{sec:dataprep}
From the original dataset, we have prepared different data setups\footnote{The arrow separates the input (left) and output (right) information. On the left part, the + symbol illustrates whether the method incorporates control attributes.}:
\begin{itemize}
  \item section $\rightarrow$ question + answer: This setup only contains the section text as input, so it serves as a baseline to compare with the subsequent setups, which consider control attributes.
  \item $\langle$EX$\rangle$ + section $\rightarrow$ question + answer: This setup considers \textit{explicitness} as a control attribute in the input.
  \item $\langle$NAR$\rangle$ + section $\rightarrow$ question + answer: This setup considers \textit{narrative} as a control attribute in the input.
  \item $\langle$NAR$\rangle$ + $\langle$EX$\rangle$ + section $\rightarrow$ question + answer: This setup considers both the explicitness and narrative attributes.
\end{itemize}

\textbf{Fair comparison}: To ensure a fair comparison in our evaluation results (Section~\ref{sec:eval_results}) between these setups, we guarantee that each section text is utilized in a single instance\footnote{A dataset instance consists of a text and corresponding ground truth question-answer pairs.} within the test set. This eliminates the creation of redundant instances, such as the repeated use of the same section text as input.

\textbf{Selection of ground truth pairs}: We ensure that all ground truth question-answer pairs within each instance refer to a single explicitness type and narrative element. This step is crucial in supporting the rationale behind Hypothesis~1 within Section~\ref{sec:eval_proced} (Evaluation Procedure).

\subsection{Implementation Details}
% few-shot details
For few-shot prompting, we use the \textit{text-davinci-003} model (GPT-3.5) from OpenAI\footnote{\url{https://platform.openai.com/docs/models/gpt-3-5}} with 
128 as the maximum token output, 0.7 for temperature and 1.0 for nucleus sampling. Following previous recommendations \citep{wang_2022_towards_aied,elkins_2023_cqg_aied}, we choose the 5-shot setting: beyond the query, 5 examples (each composed of text, question and answer) are incorporated into the prompt. The 5 examples have been randomly extracted from the train set based on the following criterion: the selected examples are consistent with the target attribute (either narrative element or explicitness) one aims to control in the generation process. So, this deliberate selection ensures a focus on a specific narrative element or explicitness (which is the goal). In the data setup where the input is just the section text, the selected examples target varied attributes.

% reference details
For the reference model, we use the pre-trained T5 encoder-decoder model \citep{raffel_2020_t5}. Firstly, T5 was trained with task-specific instructions in the form of prefixes, aligning with our methodology. Secondly, it has remarkable performance in text generation tasks, particularly in Question Generation \citep{ushio_2022_qg_t5} and CQG \citep{ghanem_2022_cqg_acl,leite_2023_cqg_aied}. Hence, we designate T5 as a smaller yet established reference model for QG and CQG. We use the \textit{t5-large} version available at Hugging Face\footnote{\url{https://huggingface.co/t5-large}}. Our maximum token input is set to 512, while the maximum token output is set to 128. During training, the models undergo a maximum of 10 epochs and incorporate early stopping with a patience of 2. Additionally, a batch size of 8 is employed. During inference, we utilize beam search with a beam width of 5. 

\section{Evaluation} \label{sec:eval}

\subsection{Evaluation Procedure} \label{sec:eval_proced}

For CQG, our evaluation protocol is based on recent work \citep{zhao_2022_cqg_acl,leite_2023_cqg_aied} that performed CQG via fine-tuning. We enrich the evaluation process with metrics related to linguistic quality \citep{wang_2022_towards_aied,elkins_2023_cqg_aied}. 

% question narrative elements control
\textbf{Narrative Control}: For assessing narrative elements control, we employ the traditional evaluation procedure in QG: directly compare the generated questions (via reference and few-shot) with the ground truth questions. Hypothesis (1) is that \textit{generated questions will be closer to the ground truth when control attributes are incorporated}. 
% Provide support to Hypothesis (1)
This hypothesis gains support from the observation that an increased closeness implies that the generated questions, prompted to match a particular narrative element, exhibit a close alignment with the ground truth questions of the same narrative element. In Figure~\ref{fig:qualitative_analyzis}, we present examples of both ground truth and generated questions that motivated this evaluation procedure, noting mainly that the beginnings of the questions are very close.
% Metrics used
To measure this closeness, we use \textit{n}-gram similarity BLEU-4 \citep{papineni_bleu_2002} and ROUGE$_L$-F1 \citep{lin_rouge_2004}. Also, for semantic similarity, we use BLEURT \citep{sellam_2020_bleurt}.
An enhancement in these metrics means that the generated questions, with a specified target narrative type, closely approximate the ground truth questions that share the same narrative type, indicating better controllability.

% question explicitness control
\textbf{Explicitness Control}: For assessing explicitness control, we resort to creating a question-answering system (QAsys) trained along FairytaleQA\footnote{We created QAsys by fine-tuning a T5 model on FairytaleQA for generating answers given the question and section text.}. The goal is to put QAsys answering questions that were generated (via reference and few-shot) and then compare QAsys answers with the answers generated (via reference and few-shot). Hypothesis (2) is that \textit{QAsys will perform significantly better on explicit than implicit generated questions}, as previously supported by FairytaleQA's authors. We also provide this evidence in Appendix~\ref{sec:appendix_qasys_gt}. To measure QAsys performance, we use ROUGE$_L$-F1 and EXACT~MATCH, a stringent scoring approach that considers a perfect match as the only acceptable outcome when comparing two strings.

% linguistic quality
\textbf{Linguistic Quality}: To evaluate the linguistic quality of the generated questions and answers, we report perplexity, grammatical error, and diversity metrics. For perplexity, our motivation is that previous studies \citep{wang_2022_towards_aied} claim there is a relation between perplexity and coherence, in a way that perplexity is inversely related to the coherence of the generated text: the lower the perplexity score, the higher the coherence. For the sake of computational efficiency, we use GPT-2 \citep{radford_2019_gpt2} to compute perplexity. For diversity, we use Distinct-3 score \citep{li_2016_diversity}, which counts the average number of distinct 3-grams in the generated text. Finally, we use Python Language Tool\footnote{\url{https://github.com/jxmorris12/language_tool_python}} to count the number of grammatical errors averaged over all generated questions and answers.

\subsection{Results and Discussion} \label{sec:eval_results}

\begin{table*}[!ht]
\footnotesize
\centering
\begin{tabular}{c|c|c|c|c|}
\cline{2-5}
 & \textbf{Data Setups} & \textbf{ROUGEL-F1 $\uparrow$} & \textbf{BLEU-4 $\uparrow$} & \textbf{BLEURT $\uparrow$} \\ \hline
\multicolumn{1}{|c|}{\multirow{4}{*}{\textbf{\begin{tabular}[c]{@{}c@{}}Reference\\Model\end{tabular}}}} & section $\rightarrow$ question + answer & 0.335 & 0.137 & 0.394 \\
\multicolumn{1}{|c|}{} & ex + section $\rightarrow$ question + answer & 0.333 & 0.138 & 0.398 \\
\multicolumn{1}{|c|}{} & nar + section $\rightarrow$ question + answer & 0.429 & \textbf{0.201} & 0.438 \\
\multicolumn{1}{|c|}{} & nar + ex + section $\rightarrow$ question + answer & \textbf{0.442} & 0.198 & 0.442 \\ \hline\hline
\multicolumn{1}{|c|}{\multirow{4}{*}{\textbf{\begin{tabular}[c]{@{}c@{}}Few-Shot\\ Prompting\end{tabular}}}} & section $\rightarrow$ question + answer & 0.339 & 0.108 & 0.397 \\
\multicolumn{1}{|c|}{} & ex + section $\rightarrow$ question + answer & 0.358 & 0.123 & 0.411 \\
\multicolumn{1}{|c|}{} & nar + section $\rightarrow$ question + answer & 0.409 & 0.168 & \textbf{0.445} \\
\multicolumn{1}{|c|}{} & nar + ex + section $\rightarrow$ question + answer & 0.402 & 0.177 & 0.441 \\ \hline
\end{tabular}
\caption{Closeness between generated and ground truth questions on the test set. All scores are 0-1.}
\label{tab:results_qg}
\end{table*}

\begin{table*}[!ht]
\footnotesize
\centering
\begin{tabular}{cc|ccc|ccc|}
\cline{3-8}
 &  & \multicolumn{3}{c|}{\textbf{ROUGEL-F1 $\uparrow$}} & \multicolumn{3}{c|}{\textbf{EXACT-MATCH $\uparrow$}} \\ \cline{2-8} 
\multicolumn{1}{c|}{} & \textbf{Data Setups} & \multicolumn{1}{c|}{Overall} & \multicolumn{1}{c|}{Explicit} & Implicit & \multicolumn{1}{c|}{Overall} & \multicolumn{1}{c|}{Explicit} & Implicit \\ \hline
\multicolumn{1}{|c|}{\multirow{2}{*}{\textbf{\begin{tabular}[c]{@{}c@{}}Reference\\Model\end{tabular}}}} & ex + section $\rightarrow$ question + answer & \multicolumn{1}{c|}{\textbf{0.661}} & \multicolumn{1}{c|}{\textbf{0.716}} & \textbf{0.513} & \multicolumn{1}{c|}{0.371} & \multicolumn{1}{c|}{0.413} & \textbf{0.259} \\
\multicolumn{1}{|c|}{} & nar + ex + section $\rightarrow$ question + answer & \multicolumn{1}{c|}{0.628} & \multicolumn{1}{c|}{0.681} & 0.487 & \multicolumn{1}{c|}{\textbf{0.383}} & \multicolumn{1}{c|}{\textbf{0.434}} & 0.250 \\ \hline\hline
\multicolumn{1}{|c|}{\multirow{2}{*}{\textbf{\begin{tabular}[c]{@{}c@{}}Few-Shot\\ Prompting\end{tabular}}}} & ex + section $\rightarrow$ question + answer & \multicolumn{1}{c|}{0.481} & \multicolumn{1}{c|}{0.531} & 0.351 & \multicolumn{1}{c|}{0.119} & \multicolumn{1}{c|}{0.143} & 0.056 \\
\multicolumn{1}{|c|}{} & nar + ex + section $\rightarrow$ question + answer & \multicolumn{1}{c|}{0.490} & \multicolumn{1}{c|}{0.556} & 0.315 & \multicolumn{1}{c|}{0.155} & \multicolumn{1}{c|}{0.185} & 0.074 \\ \hline
\end{tabular}
\caption{QAsys performance by question explicitness on the test set. All scores are 0-1.}
\label{tab:results_qasys}
\end{table*}

% Please add the following required packages to your document preamble:
% \usepackage{multirow}
\begin{table*}[!ht]
\footnotesize
\centering
\begin{tabular}{cc|ccc|ccc|}
\cline{3-8}
\multicolumn{1}{l}{} & \multicolumn{1}{l|}{} & \multicolumn{3}{c|}{\textbf{Generated Questions}} & \multicolumn{3}{c|}{\textbf{Generated Answers}} \\ \cline{2-8} 
\multicolumn{1}{l|}{} & \textbf{Data Setups} & \multicolumn{1}{c|}{\textbf{PPL $\downarrow$}} & \multicolumn{1}{c|}{\textbf{Dist-3 $\uparrow$}} & \textbf{Gram. $\downarrow$} & \multicolumn{1}{c|}{\textbf{PPL $\downarrow$}} & \multicolumn{1}{c|}{\textbf{Dist-3 $\uparrow$}} & \textbf{Gram. $\downarrow$} \\ \hline
\multicolumn{1}{|c|}{\multirow{4}{*}{\textbf{\begin{tabular}[c]{@{}c@{}}Reference\\Model\end{tabular}}}} & section $\rightarrow$ question + answer & \multicolumn{1}{c|}{197.192} & \multicolumn{1}{c|}{0.776} & 0.013 & \multicolumn{1}{c|}{303.331} & \multicolumn{1}{c|}{0.668} & 0.033 \\
\multicolumn{1}{|c|}{} & ex + section $\rightarrow$ question + answer & \multicolumn{1}{c|}{175.717} & \multicolumn{1}{c|}{0.789} & \textbf{0.005 }& \multicolumn{1}{c|}{336.649} & \multicolumn{1}{c|}{0.662} & 0.028 \\
\multicolumn{1}{|c|}{} & nar + section $\rightarrow$ question + answer & \multicolumn{1}{c|}{168.303} & \multicolumn{1}{c|}{0.782} & 0.018 & \multicolumn{1}{c|}{343.050} & \multicolumn{1}{c|}{0.597} & \textbf{0.020} \\
\multicolumn{1}{|c|}{} & nar + ex + section $\rightarrow$ question + answer & \multicolumn{1}{c|}{183.665} & \multicolumn{1}{c|}{0.789} & 0.013 & \multicolumn{1}{c|}{352.672} & \multicolumn{1}{c|}{0.560} & 0.025 \\ \hline\hline
\multicolumn{1}{|c|}{\multirow{4}{*}{\textbf{\begin{tabular}[c]{@{}c@{}}Few-Shot\\ Prompting\end{tabular}}}} & section $\rightarrow$ question + answer & \multicolumn{1}{c|}{166.160} & \multicolumn{1}{c|}{0.787} & \textbf{0.005} & \multicolumn{1}{c|}{248.966} & \multicolumn{1}{c|}{0.725} & 0.038 \\
\multicolumn{1}{|c|}{} & ex + section $\rightarrow$ question + answer & \multicolumn{1}{c|}{\textbf{143.270}} & \multicolumn{1}{c|}{0.791} & 0.013 & \multicolumn{1}{c|}{240.593} & \multicolumn{1}{c|}{\textbf{0.734}} & 0.036 \\
\multicolumn{1}{|c|}{} & nar + section $\rightarrow$ question + answer & \multicolumn{1}{c|}{155.761} & \multicolumn{1}{c|}{\textbf{0.797}} & 0.008 & \multicolumn{1}{c|}{\textbf{224.536}} & \multicolumn{1}{c|}{0.679} & \textbf{0.020} \\
\multicolumn{1}{|c|}{} & nar + ex + section $\rightarrow$ question + answer & \multicolumn{1}{c|}{153.056} & \multicolumn{1}{c|}{0.790} & 0.010 & \multicolumn{1}{c|}{260.307} & \multicolumn{1}{c|}{0.671} & 0.033 \\ \hline
\end{tabular}
\caption{Linguistic quality of generated questions and answers on the test set. Except for perplexity (PPL), all scores are 0-1.}
\label{tab:results_linguistic}
\end{table*}

\textbf{Narrative Control}: Table~\ref{tab:results_qg} presents the results using traditional QG evaluation, which directly compares the generated questions with the ground truth. Both via reference model and few-shot prompting, a significant growth in closeness to ground truth questions is observed when incorporating narrative control attributes (this happens for all metrics).
Thus, we conclude that the few-shot prompting strategy has been successfully applied to control the questions' underlying narrative elements\footnote{We provide the control results by narrative element in the Appendix~\ref{sec:cqg_per_nar}.}. 
Based on the results obtained from different metrics, we believe that the few-shot prompting might be more convenient for generating questions semantically closer to the ground truth, as reflected by the higher performance on the BLEURT metric (improves from .438 to .445). However, this improvement is not statistically significant\footnote{We perform student's t-test and find that \textit{p$_1$} $>$ .05.}. Also, the relatively lower performance on the BLEU (worsens from .201 to .168) and ROUGE$_L$-F1 (worsens from .429 to .409) metrics suggests that it may struggle with capturing certain \textit{n}-gram patterns or surface-level similarities.
These findings highlight the importance of considering multiple evaluation metrics to comprehensively understand the strengths and weaknesses of different approaches in CQG (and natural language generation in general).

\textbf{Explicitness Control}: When incorporating explicitness control attributes, as opposed to only providing section text without a specific explicitness attribute, both the reference model and few-shot prompting also result in questions closer to the ground truth\footnote{This is not true for the reference model, with ROUGE$_L$-F1.}, although less significantly than when considering the narrative attributes. To further evaluate question explicitness control, we need to analyze the question-answering results obtained by the QAsys model (as motivated earlier in Section~\ref{sec:eval_proced}).
Table~\ref{tab:results_qasys} presents the question-answering scores of QAsys when attempting to answer generated questions. For the reference model and few-shot prompting, QAsys performs significantly better on explicit than implicit generated questions, considering ROUGE$_L$-F1 and EXACT MATCH. Therefore, these results show that controlling question explicitness is possible through the few-shot strategy. It should be noted that QAsys scores are lower when answering questions generated via few-shot prompting. We strongly believe this happens because QAsys is a question-answering model trained along FairytaleQA using the T5 model, just like the reference fine-tuned models to control generation. Thus, QAsys has greater ease in answering questions generated by the reference fine-tuned models. The situation is reversed if we employ the GPT-3.5 model for QAsys (see Appendix~\ref{sec:appendix_qasys_gpt3}).

\textbf{Narrrative + Explicitness Control}: By looking at the scores in Tables~\ref{tab:results_qg} and \ref{tab:results_qasys}, when incorporating both narrative and explicitness attributes, we verify the same trend as when incorporating the attributes individually. This is true for the reference model and the few-shot prompting strategy. Therefore, using the proposed scheme, it is possible to control the generation process in few-shot prompting. While there are cases where incorporating both attributes improves the results, there are also cases where that is not the case. So, we do not find clear evidence that using multiple control attributes helps or worsens the process of controlling the generation. In Appendix~\ref{sec:appendix_nr_examples}, we show the impact of varying the number of prompt examples on the performance of few-shot prompting.

\textbf{Linguistic Quality}: Table~\ref{tab:results_linguistic} reports results for perplexity, diversity, and grammatical error to provide insight into the linguistic quality of generated questions and answers.
% perplexity
Regarding perplexity, the few-shot strategy presents questions and answers with a lower perplexity value, indicating higher coherence.
% diversity
Regarding diversity (Dist-3), the few-shot strategy presents questions with a higher diversity value than the reference model. Again, we find that the difference is not consistently statistically significant. For the answers, in contrast, we confirm that the difference is indeed consistently statistically significant.
% grammar
Finally, both methods yield an average value of zero for grammatical errors in questions and answers\footnote{We did not consider MORFOLOGIK\_RULE error from Python Language Tool, which suggests possible spelling mistakes in uncommon nouns, such as Ahtola.}. 
% Overall for linguistic
Overall, the linguistic results indicate that besides showing competence for CQG in narrative comprehension, the few-shot strategy can deliver coherent and diverse questions and answers, which motivates their use in an educational context.

\subsection{Error Analysis}

We randomly selected 105 QA pairs\footnote{15 QA pairs for each of the 7 target narrative elements.} generated from the FairytaleQA test set and analyzed potential problems. We identify 4 types of issues, which are exemplified in Figure~\ref{fig:error_analysis}. The issues found are (a) \textit{narrative misalignment} (19 in 105), (b) \textit{QA pair ambiguity} (1 in 105), (c) \textit{generic questions} (1 in 105) and (d) \textit{lengthy answers} (4 in 105).

In the first example, the generated question is not aligned with the specified ``Character'' attribute in the prompt; instead, it focuses on details about the cottage on the rock. The generated answer describes the features of the cottage but does not address the specified character (Salmon/Maie). While we are uncertain about the reasons for this model failure, it reveals the importance of verifying alignment between questions and attributes. An approach could involve implementing a post-model solution to assess alignment.

In the second example, the generated question accurately captures the attribute ``Setting'' inquiring about the location of Kittlerumpit. While the generated answer appropriately reflects the uncertainty mentioned in the text, it could be perceived as ambiguous due to the general statement (``Nobody knows''). An improvement could involve instructing the model to formulate more comprehensive questions encouraging answers that explore alternative interpretations or speculations mentioned in the text, such as the belief that Kittlerumpit stood near the Debateable Land.

In the third example, the generated question appears generic and could benefit from increased specificity. Encouraging the model to formulate questions that require detailed information, such as Ahti's dominion, possessions, and influence over the sea, may enhance the precision of the generated answers. Related to this, we find the generated answer overly lengthy. Providing guidance for more concise responses could help retain key details and shorten the answers.

\begin{figure*}[!ht]
\centering
\noindent\fbox{%
    \parbox{\textwidth}{%
    {%\scriptsize
        \begin{center}
        Example 1 (\textbf{narrative misalignment}):
        \end{center}
        \textbf{Text}: There was once a fisherman who was called Salmon (...). He had a wife called Maie; (...) In winter they dwelt in a little cottage by the shore (...). The cottage on the rock was even smaller than the other; it had a wooden bolt instead of an iron lock to the door, a stone hearth, a flagstaff, and a weather-cock on the roof...
        \newline
        \textbf{Target Narrative}: character
        \newline
        \textbf{Question}: What did the cottage on the rock have?
        \newline
        \textbf{Answer}: A wooden bolt instead of an iron lock to the door, a stone hearth, a flagstaff, and a weather-cock on the roof.
        \begin{center}
        Example 2 (\textbf{QA pair ambiguity}):
        \end{center}
        \textbf{Text}: I am going to tell you a story about a poor young widow woman, who lived in a house called Kittlerumpit, though whereabouts in Scotland the house of Kittlerumpit stood nobody knows. Some folk think that it stood in the neighbourhood of the Debateable Land...
        \newline
        \textbf{Target Narrative}: setting
        \newline
        \textbf{Question}: Whereabouts in Scotland was Kittlerumpit located?
        \newline
        \textbf{Answer}: Nobody knows.
        \begin{center}
        Example 3 (\textbf{generic questions} and \textbf{lengthy answers}):
        \end{center}
        \textbf{Text}: ``Ahti'', said they, ``is a mighty king who lives in his dominion of Ahtola, and has a rock at the bottom of the sea, and possesses besides a treasury of good things. He rules over all fish and animals of the deep; he has the finest cows and the swiftest horses that ever chewed grass at the bottom of the ocean.''...
        \newline
        \textbf{Target Narrative}: character
        \newline
        \textbf{Question}: Who is ahti?
        \newline
        \textbf{Answer}: Ahti is a mighty king who lives in his dominion of Ahtola, and has a rock at the bottom of the sea, and possesses besides a treasury of good things. He rules over all fish and animals of the deep; he has the finest cows and the swiftest horses that ever chewed grass at the bottom of the ocean.
    }
    }%
}
\caption{Examples of problematic generated question-answer pairs (error analysis) via few-shot prompting.}
\label{fig:error_analysis}
\end{figure*}

\section{Conclusion}
This work introduces a few-shot prompting strategy to address CQG for narrative comprehension, using the FairytaleQA dataset. Through experimental analysis, we observed that the generated questions, tailored  to specific attributes, closely approximate the ground truth questions of the same type. This suggests promising indicators of controllability for narrative elements and explicitness. However, our error analysis revealed instances where control did not occur, underscoring the need for further investigation when employing few-shot prompting with a state-of-the-art model like GPT-3.5.
Additionally, our findings demonstrate that the few-shot strategy can outperform the reference model in certain scenarios, however, these improvements are not consistently statistically significant.
%

% Final message
Considering our results, which align with those of a smaller yet well-established reference model for QG and CQG tasks, we find it worthwhile to employ the few-shot strategy for CQG, especially when (1) data availability is limited or (2) one favors for a ``plug-and-play'' AI-assisted approach.
% Future Work
For future work, we consider it important the application of a post-model solution to ensure that the QA pairs align with the attributes to be controlled, thereby excluding misaligned QA pairs.

\section*{Limitations}
% eval across different datasets and tasks
The effectiveness of controlling narrative elements and explicitness may vary across different datasets and tasks due to the unique characteristics of each context. While we have established that our study is focused on a specific domain and data, we recognize this limitation.
% human evaluation
Also, the lack of human evaluation is a limitation of this work. Although we believe the current evaluation process is solid for assessing the method's performance in CQG, an assessment with domain experts may help better understand the potential of CQG for educational purposes.

\section*{Acknowledgements}
This work was financially supported by Base Funding - UIDB/00027/2020 of the Artificial Intelligence and Computer Science Laboratory (LIACC) funded by national funds through FCT/MCTES (PIDDAC). Bernardo Leite is supported by a PhD studentship (reference 2021.05432.BD), funded by Funda\c{c}\~{a}o para a Ci\^{e}ncia e a Tecnologia (FCT).

% Entries for the entire Anthology, followed by custom entries
\bibliographystyle{apalike}
{\small\bibliography{example}}

\appendix

\section{Analyzing QAsys Performance in ground truth Questions} \label{sec:appendix_qasys_gt}

In Section~\ref{sec:eval_proced}, we hypothesize that QAsys, a question-answering system, will perform significantly better on explicit questions compared to implicit ones. Supporting this hypothesis is important for validating our evaluation procedure regarding question explicitness control. So, in Table~\ref{tab:qasys_gt}, we present QAsys's results on ground truth questions (test set). As shown in the results, QAsys performs significantly better on explicit questions, providing support for the hypothesis.

\section{Controllability by Nar. Element} \label{sec:cqg_per_nar}

To gain a more comprehensive understanding of narrative element control, Table~\ref{tab:results_qg_per_nar} provides a breakdown of results for each narrative element. In the case of few-shot prompting, we observe that transitioning from using the section text solely as input to incorporating \textsc{$<$nar$>$} yields the most substantial improvement (0.227) in the ``Feeling'' narrative element. On the other, the reference model demonstrates the best enhancement of 0.271 in the ``Setting'' narrative element, with the second best improvement in the ``Feeling'' element as well (0.217).

\section{Analyzing QAsys Performance via Few-Shot Prompting} \label{sec:appendix_qasys_gpt3}

% remind the reader
In Section~\ref{sec:eval_results}, where we evaluate explicitness control, we notice that QAsys performance improves when answering questions generated by the reference models. This observation led us to speculate that this advantage stems from the fact that QAsys, like the reference models to control question generation, is also trained using the same T5 model.
% what we did
To further analyse this, we conducted an additional experiment where we replaced the T5 model with GPT-3.5 to answer the generated questions.

% show results
Table~\ref{tab:results_qasys_gpt3} shows the QAsys scores, obtained through the few-shot strategy with GPT-3.5, when attempting to respond to generated questions.
% explain what happened
We now observe a reversal in the situation, with QAsys achieving higher scores when addressing questions generated via few-shot prompting and lower scores for those generated from the reference model. This supports our previous speculation, suggesting that a question-answering model yields improved results when tasked with answering questions generated by a model with the same architecture.
% reinforce conclusion
Most importantly, these results reinforce our main conclusion: regardless of the model, QAsys performs better when dealing with explicit as opposed to implicit generated questions, demonstrating that both a few-shot strategy and reference model (via fine-tuning) effectively enable the control of question explicitness.

\section{Varying the Number of Prompt Examples}
\label{sec:appendix_nr_examples}

\begin{figure}[htp]
    \centering
    \includegraphics[scale=0.25]{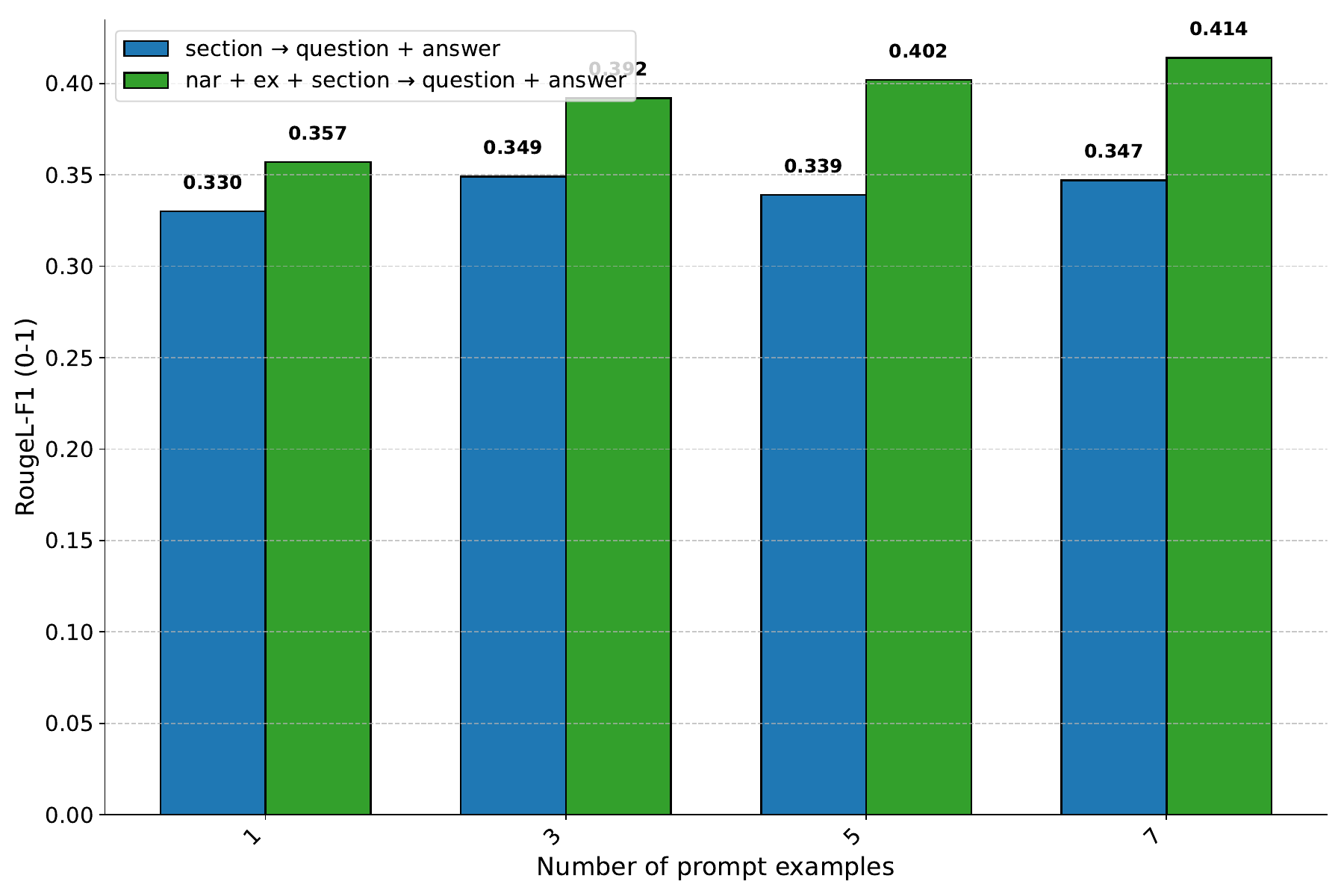}
    \caption{Results of varying the number of examples in few-shot prompting for question narrative control.} 
    \label{fig:ablation_1}
\end{figure}

While our current approach uses 5 examples, aligned with previous recommendations \citep{wang_2022_towards_aied,elkins_2023_cqg_aied}, we explore alternative numbers of examples. Figure~\ref{fig:ablation_1} shows the impact on closeness results (for assessing narrative elements control) when using 1, 3, 5 and 7 prompt examples. Our conclusions are as follows:
\begin{itemize}
    \item In line with our primary results (using 5 prompt examples), we consistently observe a significant increase in closeness when incorporating narrative control attributes, regardless of the number of prompt examples.
    \item With the inclusion of control attributes (see green bars), increasing the number of prompt examples leads to improved narrative closeness results.
    \item In the absence of control attributes (see blue bars), increasing the number of prompt examples does not yield a consistent improvement in closeness results. This is expected, as the goal in this scenario is not to provide specific prompt examples for controlling a particular type of question. Instead, the goal is to generate questions that do not target specific attributes, relying on prompt examples that address various narrative and explicitness attributes.
\end{itemize}

\begin{figure}[htp]
    \centering
    \includegraphics[scale=0.25]{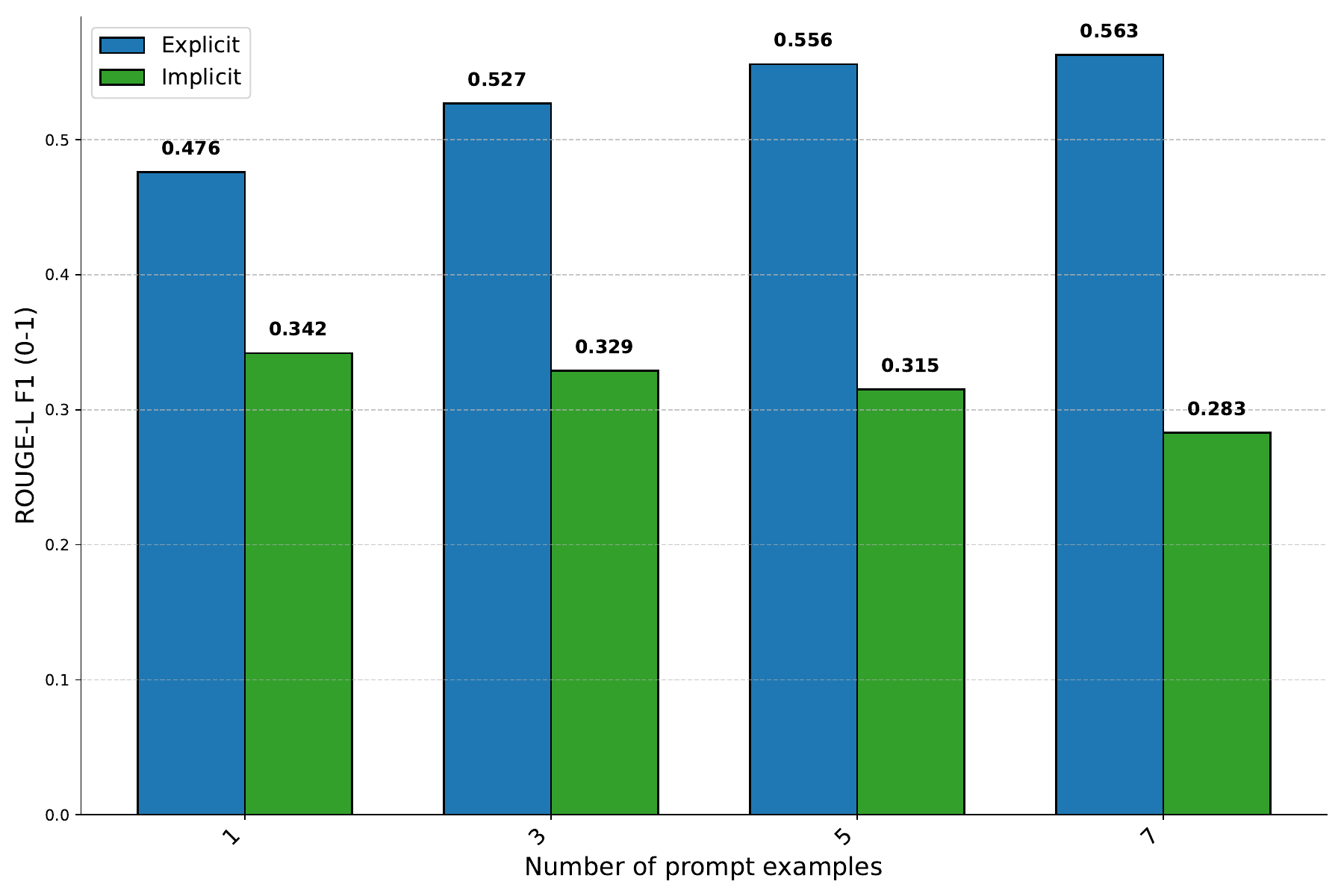}
    \caption{Results of varying the number of examples in few-shot prompting for question explicitness control.} 
    \label{fig:ablation_2}
\end{figure}

Figure~\ref{fig:ablation_2} presents the question-answering scores of QAsys when attempting to answer generated questions (for assessing explicitness control), which were generated by experimenting with different numbers of prompt examples. We conclude the following:
\begin{itemize}
    \item Consistent with our main results (where we use only 5 prompt examples), QAsys consistently outperforms on explicit questions compared to implicit ones, regardless of the number of prompt examples.
    \item As the number of prompt examples increases, QAsys consistently improves for explicit questions and consistently underperforms for implicit questions. We posit that this can be related to the larger set of examples ``helping'' the model in refining both explicit and implicit question generation.
    \item As the number of prompt examples increases, the gap in QAsys results widens between explicit and implicit questions. This indicates the advantage of increasing the number of prompt examples for better explicitness control.
\end{itemize}

\begin{table*}[!ht]
\footnotesize
\centering
\begin{tabular}{|ccc|ccc|}
\hline
\multicolumn{3}{|c|}{\textbf{ROUGEL-F1}}                                 & \multicolumn{3}{c|}{\textbf{EXACT-MATCH}}                               \\ \hline
\multicolumn{1}{|c|}{Overall} & \multicolumn{1}{c|}{Explicit} & Implicit & \multicolumn{1}{c|}{Overall} & \multicolumn{1}{c|}{Explicit} & Implicit \\ \hline
\multicolumn{1}{|c|}{0.597}   & \multicolumn{1}{c|}{0.732}    & 0.194    & \multicolumn{1}{c|}{0.315}   & \multicolumn{1}{c|}{0.403}    & 0.051    \\ \hline
\end{tabular}
\caption{QAsys results in ground truth explicit and implicit questions on the test set. All scores are 0-1.}
\label{tab:qasys_gt}
\end{table*}

\begin{table*}[!ht]
\footnotesize
\centering
\begin{tabular}{cc|ccccccc|}
\cline{3-9}
\multicolumn{1}{l}{}                                                                                         & \multicolumn{1}{l|}{}                              & \multicolumn{7}{c|}{\textbf{Narrative Elements}}                                                                                                                                           \\ \cline{2-9} 
\multicolumn{1}{c|}{}                                                                                        & \textbf{Data Setups}                               & \multicolumn{1}{c|}{Chara.} & \multicolumn{1}{c|}{Setting} & \multicolumn{1}{c|}{Action} & \multicolumn{1}{c|}{Feeling} & \multicolumn{1}{c|}{Causal} & \multicolumn{1}{c|}{Out.}  & Pred. \\ \hline
\multicolumn{1}{|c|}{\multirow{3}{*}{\textbf{\begin{tabular}[c]{@{}c@{}}Reference\\Model\end{tabular}}}}       & section $\rightarrow$ question + answer            & \multicolumn{1}{c|}{0.320}  & \multicolumn{1}{c|}{0.279}   & \multicolumn{1}{c|}{0.372}  & \multicolumn{1}{c|}{0.300}   & \multicolumn{1}{c|}{0.381}  & \multicolumn{1}{c|}{0.273} & 0.240 \\
\multicolumn{1}{|c|}{}                                                                                       & nar + section $\rightarrow$ question + answer      & \multicolumn{1}{c|}{\textbf{0.360}}  & \multicolumn{1}{c|}{0.550}   & \multicolumn{1}{c|}{0.461}  & \multicolumn{1}{c|}{0.517}   & \multicolumn{1}{c|}{0.409}  & \multicolumn{1}{c|}{0.374} & 0.379 \\
\multicolumn{1}{|c|}{}                                                                                       & nar + ex + section $\rightarrow$ question + answer & \multicolumn{1}{c|}{0.350}  & \multicolumn{1}{c|}{\textbf{0.615}}   & \multicolumn{1}{c|}{0.461}  & \multicolumn{1}{c|}{\textbf{0.568}}   & \multicolumn{1}{c|}{\textbf{0.419}}  & \multicolumn{1}{c|}{\textbf{0.447}} & \textbf{0.450} \\ \hline\hline
\multicolumn{1}{|c|}{\multirow{3}{*}{\textbf{\begin{tabular}[c]{@{}c@{}}Few-Shot\\ Prompting\end{tabular}}}} & section $\rightarrow$ question + answer            & \multicolumn{1}{c|}{0.254}  & \multicolumn{1}{c|}{0.307}   & \multicolumn{1}{c|}{0.449}  & \multicolumn{1}{c|}{0.305}   & \multicolumn{1}{c|}{0.303}  & \multicolumn{1}{c|}{0.324} & 0.300 \\
\multicolumn{1}{|c|}{}                                                                                       & nar + section $\rightarrow$ question + answer      & \multicolumn{1}{c|}{0.277}  & \multicolumn{1}{c|}{0.380}   & \multicolumn{1}{c|}{0.496}  & \multicolumn{1}{c|}{0.532}   & \multicolumn{1}{c|}{0.377}  & \multicolumn{1}{c|}{0.387} & 0.335 \\
\multicolumn{1}{|c|}{}                                                                                       & nar + ex + section $\rightarrow$ question + answer & \multicolumn{1}{c|}{0.296}  & \multicolumn{1}{c|}{0.365}   & \multicolumn{1}{c|}{\textbf{0.498}}  & \multicolumn{1}{c|}{0.516}   & \multicolumn{1}{c|}{0.367}  & \multicolumn{1}{c|}{0.337} & 0.327 \\ \hline
\end{tabular}
\caption{Closeness (ROUGE$_L$-F1 $\uparrow$) between generated and ground truth questions on the test set by narrative element. All scores are 0-1.}
\label{tab:results_qg_per_nar}
\end{table*}

\begin{table*}[!ht]
\footnotesize
\centering
\begin{tabular}{cc|ccc|ccc|}
\cline{3-8}
                                                                                                             &                                                    & \multicolumn{3}{c|}{\textbf{ROUGEL-F1 $\uparrow$}}                                 & \multicolumn{3}{c|}{\textbf{EXACT-MATCH $\uparrow$}}                               \\ \cline{2-8} 
\multicolumn{1}{c|}{}                                                                                        & \textbf{Data Setups}                               & \multicolumn{1}{c|}{Overall} & \multicolumn{1}{c|}{Explicit} & Implicit & \multicolumn{1}{c|}{Overall} & \multicolumn{1}{c|}{Explicit} & Implicit \\ \hline
\multicolumn{1}{|c|}{\multirow{2}{*}{\textbf{\begin{tabular}[c]{@{}c@{}}Reference\\Model\end{tabular}}}}       & ex + section $\rightarrow$ question + answer       & \multicolumn{1}{c|}{0.517}   & \multicolumn{1}{c|}{0.580}    & 0.352    & \multicolumn{1}{c|}{0.160}   & \multicolumn{1}{c|}{0.185}    & 0.093    \\
\multicolumn{1}{|c|}{}                                                                                       & nar + ex + section $\rightarrow$ question + answer & \multicolumn{1}{c|}{0.479}   & \multicolumn{1}{c|}{0.522}    & 0.365    & \multicolumn{1}{c|}{0.193}   & \multicolumn{1}{c|}{0.224}    & 0.111    \\ \hline\hline
\multicolumn{1}{|c|}{\multirow{2}{*}{\textbf{\begin{tabular}[c]{@{}c@{}}Few-Shot\\ Prompting\end{tabular}}}} & ex + section $\rightarrow$ question + answer       & \multicolumn{1}{c|}{\textbf{0.754}}   & \multicolumn{1}{c|}{\textbf{0.785}}    & \textbf{0.673}    & \multicolumn{1}{c|}{\textbf{0.325}}   & \multicolumn{1}{c|}{\textbf{0.360}}    & \textbf{0.231}    \\
\multicolumn{1}{|c|}{}                                                                                       & nar + ex + section $\rightarrow$ question + answer & \multicolumn{1}{c|}{0.674}   & \multicolumn{1}{c|}{0.727}    & 0.532    & \multicolumn{1}{c|}{0.256}   & \multicolumn{1}{c|}{0.294}    & 0.157    \\ \hline
\end{tabular}
\caption{QAsys performance (via few-shot) by question explicitness on the test set. All scores are 0-1.}
\label{tab:results_qasys_gpt3}
\end{table*}

\begin{figure*}[!ht]
\centering
\noindent\fbox{%
    \parbox{\textwidth}{%
    {%\scriptsize
        \begin{center}
        Question-Answer Pairs (\textbf{character}): %character
        \end{center}
        \textbf{GT}: Who wished the princesses into the earth? Three trolls.
        \newline
        \textbf{GEN}: Who told the youth he could have whichever one of them he wanted? The third princess.
        \newline
        \textbf{GT}: Who was a tall, handsome man, with dark hair, and eyes like sloes? Paul.
        \newline
        \textbf{GEN}: Who was Snorro's bosom friend and companion? An enormous grey-headed Raven.
    
        \begin{center}
        Question-Answer Pairs (\textbf{setting}): % setting
        \end{center}
        \textbf{GT}: Where did the bannock escape to after running away from the farmhouse? A little cottage standing among the heather.
        \newline
        \textbf{GEN}: Where was the little cottage located? Among the heather.
        \newline
        \textbf{GT}: Where was the small village? The foot of some high mountains.
        \newline
        \textbf{GEN}: Where was the village located? At the foot of some high mountains.
        
        \begin{center}
        Question-Answer Pairs (\textbf{action}): % action
        \end{center}
        \textbf{GT}: What did Matte and Maie do on Saturdays? Sailed to the nearest town, sold their fish, and went to church on Sunday.
        \newline
        \textbf{GEN}: What did Maie and Matte do to provide for themselves? They fished for salmon in spring, for herring in summer, and for cod in winter.
        \newline
        \textbf{GT}: What did the cow do in the autumn? Went to sea.
        \newline
        \textbf{GEN}: What did the cow do to find food? She went out and cropped the seaweed which grew in great abundance near the shore.

        \begin{center}
        Question-Answer Pairs (\textbf{feeling}): % feeling
        \end{center}
        \textbf{GT}: How did the princess feel in her new home? Happy.
        \newline
        \textbf{GEN}: How did the princess feel when her mother visited her in her new home? Welcomed and happy.
        \newline
        \textbf{GT}: How did Ryn Jin feel when the Happy Hunter asked to leave? Sorrow.
        \newline
        \textbf{GEN}: How did King Ryn Jin feel when he had to part with Mikoto? Overcome with sorrow.
    }
    }%
}
\caption{Examples of both ground truth (\textbf{GT}) and generated (\textbf{GEN}) questions, highlighting the close similarity, mainly in the beginnings of the questions. This observation has inspired the procedure used in the narrative control evaluation (Section \ref{sec:eval_proced}).}
\label{fig:qualitative_analyzis}
\end{figure*}

\begin{figure*}[!ht]
\centering
\noindent\fbox{%
    \parbox{\textwidth}{%
    {%\scriptsize
        Generate questions and answers targeting the following narrative element: outcome resolution
        \newline\newline
        %\textbf{Text}: The boy went out again and brought home the little animal, which he asked his grandfather to boil, that they might feast on it. He humored the boy in this, and he encouraged him to go on in acquiring the knowledge of hunting, until he could kill deer and the larger kinds of game; and he became, as he grew up, an expert hunter. As they lived alone, and away from other Indians, the curiosity of the stripling was excited to know what was passing in the world. One day he came to the edge of a prairie, where he saw ashes like those at his grandfather's lodge, and lodge-poles left standing.
        %\newline
        %\textbf{Question}: What happened after the grandfather told the boy about the rabbit?
        %\newline
        %\textbf{Answer}: The boy went out again and brought home the little animal, which he asked his grandfather to boil, that they might feast on it.
        %\newline
        %\newline
        \textbf{Text}: One day Isaac had put out a few miles to sea to fish, when suddenly a dark fog fell. In a flash such a tremendous storm broke, that he had to throw all his fish overboard in order to lighten ship and save his life...
        %Even then it was very hard to keep the boat afloat. He steered a careful course between and across the mountainous waves, which seemed ready to swallow him from moment to moment. After he had kept on for five or six hours in this manner, he thought that he ought to touch land somewhere. But time went by, and the storm and fog grew worse and worse. Then he began to realize that either he was steering out to sea, or that the wind had veered, and at last he made sure the latter was the case. He sailed on and on without a sight of land.
        \newline
        \textbf{Question}: What happened because of the tremendous storm?
        \newline
        \textbf{Answer}: Isaac had to throw all his fish overboard in order to lighten ship and save his life.
        \newline
        \newline
        \textbf{Text}: 
        %The son had often heard tell of her, and one fall, when his parents had already come home from the mountain pasture, he put on his full uniform, saddled his service horse, thrust his pistols in the holsters, and thus rode up into the hills. 
        When he rode toward the pasture, such a fire burned in the herdsman's hut that it lit up every road, and then he knew that the mountain folk were inside...
        %So he tied his horse to a pine-tree, took a pistol from its holster, crept up to the hut, and peeped through the window. And there sat an old man and a woman who were quite crooked and shriveled up with age, and so unspeakably ugly that he had never seen anything like it in his life; but with them was a maiden, and she was so surpassingly beautiful that he fell in love with her at once, and felt that he could not live without her. All had cow's tails, and the lovely maiden, too. And he could see that they had only just arrived, for everything was in disorder. The maiden was busy washing the ugly old man, and the woman was building a fire under the great cheese-kettle on the hearth.
        \newline
        \textbf{Question}: What happened because such a fire burned in the herdsman's hut?
        \newline
        \textbf{Answer}: It lit up every road.
        \newline
        \newline
        (...+3 prompt examples...)
        \newline
        \newline
        \textbf{Text}:
        %But this clever son replied: 'If I give you my cake and wine I shall have none left for myself; you just go your own way;' and he left the little man standing there and went further on into the forest. There he began to cut down a tree, but before long he made a false stroke with his axe, and cut his own arm so badly that he was obliged to go home and have it bound up. 
        %Then the second son went to the forest, and his mother gave him a good cake and a bottle of wine as she had to his elder brother. He too met the little old grey man, who begged him for a morsel of cake and a draught of wine. 
        But the second son spoke most sensibly too, and said: 'Whatever I give to you I deprive myself of. Just go your own way, will you?' Not long after his punishment overtook him, for no sooner had he struck a couple of blows on a tree with his axe, than he cut his leg so badly that he had to be carried home.
        \newline
        \textbf{Question}: \textcolor{blue}{\textit{What happened to the second son?}}
        \newline
        \textcolor{blue}{\textbf{\textit{Answer}}: \textit{He cut his leg so badly that he had to be carried home.}}

    }
    }%
}
\caption{Example of CQG targeting the following narrative element: outcome resolution. Generated text is shown in italics and blue. Texts are from the FairytaleQA dataset.}
\label{fig:cqg_nar}
\end{figure*}

\begin{figure*}[!ht]
\centering
\noindent\fbox{%
    \parbox{\textwidth}{%
    {%\scriptsize

        Generate implicit questions and answers
        \newline\newline
        \textbf{Text}: 
        %It was the rainy season at Nara, and floods were reported every day as doing damage in the neighborhood. The river Tatsuta, which flowed through the Imperial Palace grounds, was swollen to the top of its banks. 
        The roaring of the torrents of water rushing along a narrow bed so disturbed the Emperor's rest day and night, that a serious nervous disorder was the result...
        %An Imperial Edict was sent forth to all the Buddhist temples commanding the priests to offer up continuous prayers to Heaven to stop the noise of the flood. But this was of no avail." Strange indeed it seemed to all those standing round. 
        The waters ceased their roaring, and the river was quiet in direct answer to her prayer. After this the Emperor soon recovered his health...
        %His Majesty was highly pleased, and sent for her to the Palace and rewarded her with the rank of Chinjo-that of Lieutenant-General-to distinguish her. From that time she was called Chinjo-hime, or the Lieutenant-General Princess, and respected and loved by all.
        \newline
        \textbf{Question}: How did the ceasing of the water roar allow the Emperor to recover in his health?
        \newline
        \textbf{Answer}: He could now sleep soundly.
        \newline
        \newline
        \textbf{Text}: That one thing was that there was one room in the Castle--a room which stood at the end of a passage by itself--which she could never enter, as her husband always carried the key...
        %And as, when she asked him the reason of this, he always made an excuse of some kind. She made up her mind that she would not seem as if she did not trust him, so she asked no more questions about the matter. 
        But one day the Prince chanced to leave the door unlocked. As he had never told her not to do so, she went in. There she saw Princess Gold-Tree lying on the silken couch, looking as if she were asleep... 
        %"Is she dead, or is she only sleeping?" she said to herself. She went up to the couch and looked closely at the Princess. And there, sticking in her little finger, she discovered a curiously shaped needle. "There hath been evil work here," she thought to herself. "If that needle be not poisoned, then I know naught of medicine." And, being skilled in leechcraft, she drew it carefully out.
        \newline
        \textbf{Question}: Why wasn't the second wife allowed to enter one room in the Castle?
        \newline
        \textbf{Answer}: Gold-Tree was in the room.
        %\newline
        %\newline
        %\textbf{Text}: All would have gone well if Farquhar had only looked where he was going. He did not, being deeply engaged in making love to a young Fairy Maiden by his side, so he never saw a cottage that was standing right in his way. He struck against the chimney and stuck fast in the thatch. His companions sped merrily on, not noticing what had befallen him. He was left to disentangle himself as best he could. As he was doing so he chanced to glance down the wide chimney. In the cottage kitchen he saw a comely young woman dandling a rosy-cheeked baby.
        %\newline
        %\textbf{Question}: Why did Farquhar strike against the chimney and stuck fast in the thatch?
        %\newline
        %\textbf{Answer}: He did not look where he was going.
        \newline
        \newline
        (...+3 prompt examples...)
        \newline
        \newline
        \textbf{Text}: 'Oh, only the words of an old rhyme that keeps running in my head', answered the old woman; and she raised her voice and went on: Oh, Ahti, with the long, long beard, Who dwellest in the deep blue sea, A thousand cows are in thy herd, I pray thee give one onto me. 'That's a stupid sort of song' said Matte...
        %'What else should one beg of the sea-king but fish? But such songs are not for Sunday.' His wife pretended not to hear him, and sang and sang the same tune all the time they were on the water. Matte heard nothing more as he sat and rowed the heavy boat, while thinking of his cracked pipe and the fine tobacco. 
        Then they returned to the island, and soon after went to bed. But neither Matte nor Maie could sleep a wink; the one thought of how he had profaned Sunday, and the other of Ahti's cow.
        \newline
        \textbf{Question}: \textcolor{blue}{\textit{What did Maie keep singing during their journey?}}
        \newline
        \textcolor{blue}{\textbf{\textit{Answer}}: \textit{An old rhyme about Ahti and his thousand cows.}}

    }
    }%
}
\caption{Example of CQG targeting implicit questions. Generated text is shown in italics and blue. Texts are from the FairytaleQA dataset.}
\label{fig:cqg_ex}
\end{figure*}

\end{document}